\def\year{2020}\relax
\documentclass[letterpaper]{article} 
\usepackage{aaai20}  
\usepackage{times}  
\usepackage{helvet} 
\usepackage{courier}  
\usepackage[hyphens]{url}  
\usepackage{graphicx} 
\usepackage{graphicx}  
\usepackage{amsmath,amsfonts,lipsum,amssymb}
\usepackage{comment}
\usepackage{tabularx,booktabs}
\title{Fragmentation Coagulation Based Mixed Membership Stochastic Blockmodel}

\author{Zheng Yu,\textsuperscript{\rm 1} Xuhui Fan,\textsuperscript{\rm 2} Marcin Pietrasik,\textsuperscript{\rm 1} Marek Reformat\textsuperscript{\rm 1, \rm 3}
 \\ 
\textsuperscript{\rm 1}Department of Electrical and Computer Engineering, University of Alberta\\ 
\textsuperscript{\rm 2}School of Mathematics \& Statistics,
University of New South Wales\\
\textsuperscript{\rm 3}Information Technology Institute, University of Social Sciences, Poland\\
zy3@ualberta.ca, xuhui.fan@unsw.edu.au, pietrasi@ualberta.ca, reformat@ualberta.ca
}
\begin{document}

\maketitle

\begin{abstract}
The Mixed-Membership Stochastic Blockmodel~(MMSB) is proposed as one of the state-of-the-art Bayesian relational methods suitable for learning the complex hidden structure underlying the network data. However, the current formulation of MMSB suffers from the following two issues: (1), the prior information~(e.g. entities' community structural information) can not be well embedded in the modelling; (2), community evolution can not be well described in the literature. Therefore, we propose a non-parametric fragmentation coagulation based Mixed Membership Stochastic Blockmodel (fcMMSB). Our model performs entity-based clustering to capture the community information for entities and linkage-based clustering to derive the group information for links simultaneously. Besides, the proposed model infers the network structure and models community evolution, manifested by appearances and disappearances of communities, using the discrete fragmentation coagulation process (DFCP). By integrating the community structure with the group compatibility matrix  we derive a generalized version of MMSB. An efficient Gibbs sampling scheme with Polya Gamma (PG) approach is implemented for posterior inference. We validate our model on synthetic and real world data.
\end{abstract}

\section{Introduction}

\begin{figure*}

    \centering
    \includegraphics[width=0.95\textwidth]{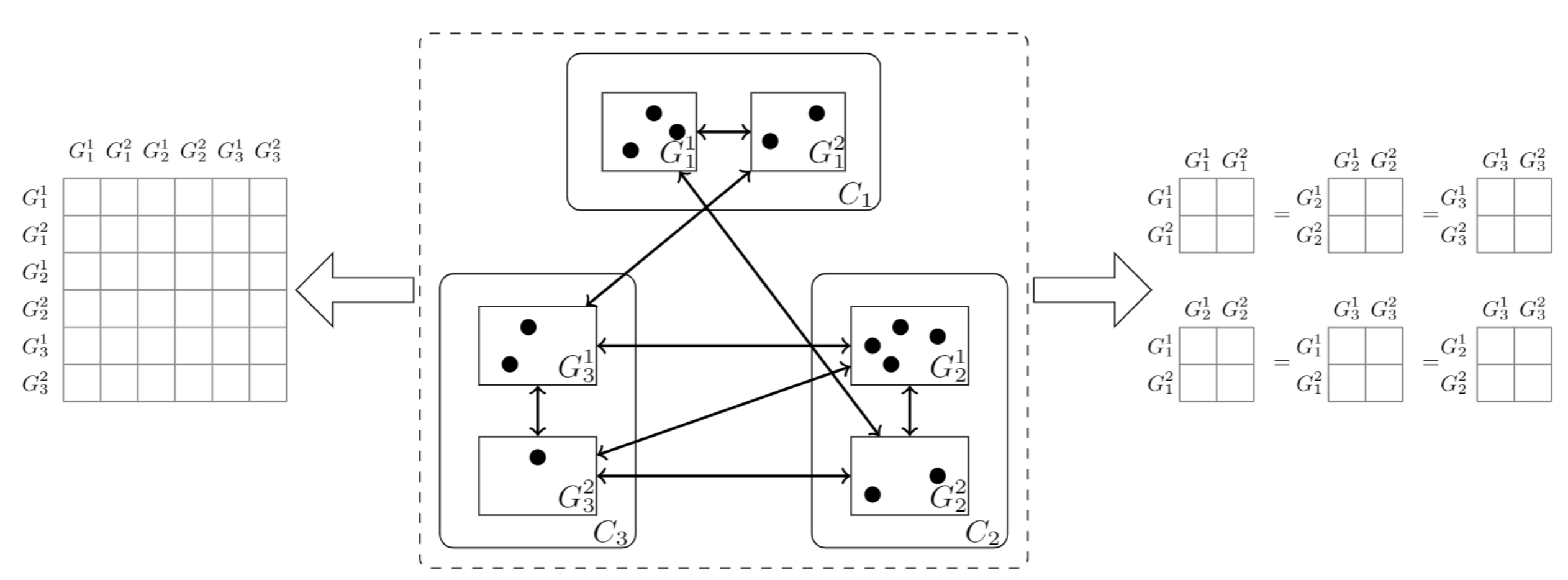}
\caption{An example to illustrate the intuition of the proposed model. Each community $(C_1 ,C_2$ and $C_3)$ consists of two groups $G^1$ and $G^2$. Entities within each group are represented by black dots. Four types of interactions are considered: within/across groups and within/across communities. In MMSB, a $6 \times 6$ compatibility matrix can be used (left part). In our model, it is represented by 2 compatibility matrices: one representing the group relations within communities and another representing the group relations across communities. (right part)}
\label{fig:toyexample}
\end{figure*}

Analysis of complex networks is an important research topic leading to a variety of useful applications. 
To this end, many interesting and promising approaches have been proposed to address various challenges in investigating these complex networks. The Mixed-Membership Stochastic Blockmodel~(MMSB) \cite{airoldi2008mixed} is one such state-of-the-art model in using Bayesian methods to discover meaningful underlying hidden structure. In general, MMSB assumes each entity in the network has a mixed-membership distribution over the groups. To generate the link between two entities, each entity would sample a belonging group from its mixed-membership distribution. The compatibility value between these two sampled groups would then determine the probability of generating this link.

MMSB has garnered considerable interest in recent years, however, it is not good at embedding certain prior information such as, for instance, the entities' community structure. When the entities in the network are assumed to have a mixed-membership distribution over the groups, the entity itself would belong to only one community. That is to say, we should consider two types of clustering in MMSB: entity-based clustering (i.e. communities for entities) and linkage-based clustering (i.e. groups for links.)

For example, each footballer can play multiple positions (groups) in one match while only belonging to one team (community). This situation is quite common in the real world.  Besides,  consider the more general example in Figure \ref{fig:toyexample} where there are three communities $\{C_i| i\in{1,2,3}\}$ in the network, each composed of two groups $(G_i^1,G_i^2)$. If we use a $6 \times 6$ compatibility matrix, this will hinder interpretability because entities that should belong to groups in the same community may belong to groups in different communities. Under this setting, the MMSB  can't not infer any community information about entities. Moreover, the size of the compatibility matrix is bigger than the true one (or the proposed one in Figure \ref{fig:toyexample}.) which may lead to an overfitting problem.

Furthermore, another issue will also be prominent under the dynamic setting. Recall that with respect to temporal dynamics, most of MMSB-based temporal models focus on correlation among groups in the adjacent time slice. However, the size of their compatibility matrices is same across time which leads to another shortcoming. Consider, for instance, a simple case where there is a complex network with just 2 time slices. At time slice 1, there is one community that consists of 4 groups. It is reasonable to use MMSB with a $4 \times 4$ compatibility matrix to represent it. However, at time slice 2, the community splits into two communities. Each community still consists of 4 groups but the entities originally in the same group may have different relations based on the community they belong to. Thus a compatibility matrix of size $8 \times 8$ is more suitable at time slice 2. This causes a problem when selecting the compatibility matrix size in the MMSB. Choosing the $4 \times 4$ matrix will lead to an underfitting problem while choosing the $8\times8$ one will lead to an overfitting problem. 

In this work, we focus on the following problems: 
\begin{itemize}
\item In a complex network, we should consider two types of clustering: entity-based clustering (communities for entities); and linkage-based clustering (groups for links). MMSB-based models only adapt the second one in both static and dynamic setting and this will hinder community interpretability. 
\item Community evolution exists in complex networks across time. MMSB-based models are not able to capture these changes by merely adjusting the size of the compatibility matrix as they use a fixed size compatibility matrix across time. 
\end{itemize}

To handle these two problems, we propose the fragmentation coagulation based Mixed Membership Stochastic Blockmodel (fcMMSB). 

To enrich the structure of MMSB, we introduce a community level to MMSB in which the Chinese restaurant process (CRP) is used to partition entities. Due to the nonparametric property of CPR, the number of communities doesn't need to be specified and this  makes the model more flexible. For entities in the same community, MMSB is carried out independently to enable each entity to hold multiple groups. 

To distinguish the group relations within/across communities, we make use of two compatibility matrices, one for modeling relations between groups in the same community and one for modeling relations between groups in different communities. Specifically, we introduce an across community adjustment parameter which acts as a modifier on the intra group relations across communities so that intra group relations are different if the groups belong to different communities.

Furthermore, to handle the issue in the dynamic setting, we incorporate the discrete fragmentation coagulation process (DFCP) \cite{elliott2012scalable,luo2017tracking} to model the community evolution across time. This allows us to release the limitation of the fixed size compatibility matrix in MMSB across time. The reason is that DFCP can automatically learn the number of communities at each time slice. Also, the changes in the number of communities would influence the entities' group membership. Therefore, this will influence the size of compatibility matrix implicitly. Besides, DFCP helps to model situations such as community splitting and merging while also generalizing MMSB such that when there is only one community in the network, it just turns back to the vanilla MMSB. With this approach, communities can merge into super communities or split into small communities.

\section{Model Formulation}
In fcMMSB, our task is to do link prediction for the unobserved entity interactions, based on the observed ones. We focus on binary-valued interaction with a total number of $N$ entities at $T$ time slices. Formally, these interactions can be defined as a binary $3$-d tensor $\mathbf{X} \in \{{0,1\}}_{N \times N}^{T}$, where $x_{ij}^t=1$ represents a directed interaction between entity $u_i$ and entity $u_j$ at time slice $t$, and $x_{ij}^t=0$ represents no interaction. Other format of the observed interactions is possible by considering different forms of the likelihood functions.

\subsection{Modelling Community Evolution Using DFCP}
In our model, each entity (individual) is associated with a community, so community evolution influences relations between entities. Consider, for example, a scenario where corporations are communities, the branches within these corporations (IT, accounting, etc.) are groups, and the network models relations between employees. In the case of a corporate merger, the interactions between employees in the same branches of the merging corporations will increase. In general, we can categorize community evolution into four types: appearance, disappearance, split, and merge. We use fragmentation and coagulation to depict all four types of changes such that coagulation and fragmentation correspond to merging and splitting, respectively. Community appearance and disappearance can be viewed as extensions of community splits and merges. Since communities evolve, it is hard to know the number of communities a priori, thus our model infers the number of communities  using non-parametric Bayes.

\begin{figure}
\centering
\includegraphics[width=.95\columnwidth]{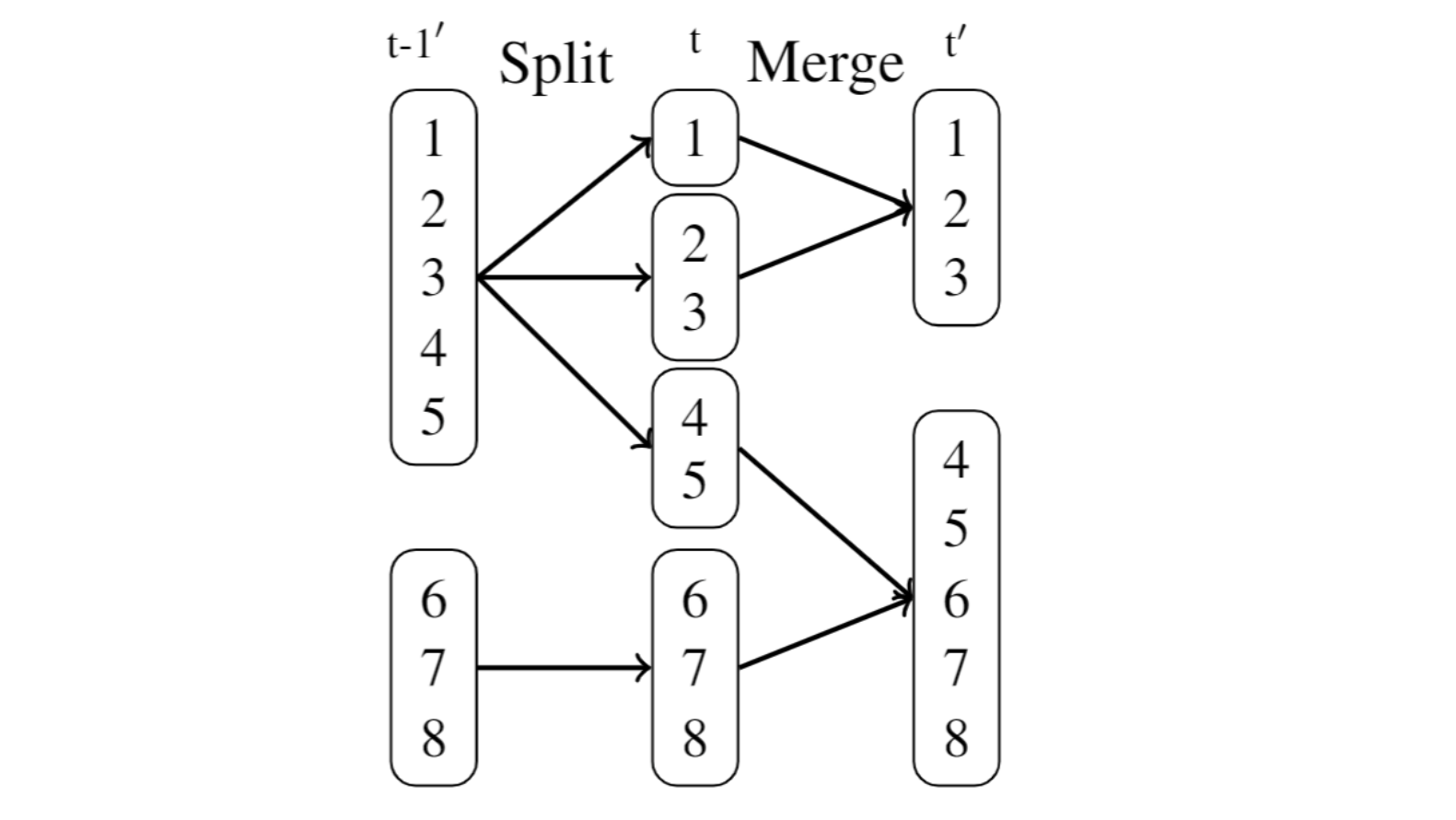}
\caption{Visualization of fragmentation and coagulation processes in fcMMSB. For example, the community of $\{1,2,3,4,5\}$ at time ${t-1}'$ will first be split into $3$ small sub-communities $\{1\}, \{2, 3\}, \{4, 5\}$ and then be re-clustered into communities at time $t'$.} 
\label{visual of fc}
\end{figure}

We adopt the DFCP framework to implement these two operations. DFCP is a non-parametric dynamic clustering process where clusters are first split (fragmentation) and then merged (coagulation). DFCP performs the fragmentation and coagulation processes alternately. 
To describe the procedures of fragmentation and coagulation, we define a set of disjoint non-empty subsets, $\nu^t=\{\chi_1^t,...,\chi_r^t\}$ where $\chi_h^t$ is a latent community $h$ at $t$ and $r$ is the number of communities at time $t$. Furthermore, each subset $\chi_h^t$ consists of disjoint entities $u_i$ in the network. Figure \ref{visual of fc} provides the visualization of fragmentation and coagulation processes. In our model, we process fragmentation and coagulation at times $t-1'$ and $t$, respectively. At time $t-1'$, the fragmentation process partitions each community $\chi_h^{t-1'}$ from $\nu^{t-1'}$ while at time $t$ the obtained partitions are coagulated into a new set  of communities $\nu^{t'} =\{\chi_1^{t'},...,\chi_r^{t'}\}$.

Now, we provide the generative process for communities using DFCP. To sample community indicator $z_i^t$ for each entity $u_i$ where $i \in \{1,...,N\}$, we start an initialization with CRP at $t=0$ as:

$\text{Init}({z_i}^{t}):\text{p}({z_i}^{0} = h|\mathbf{z}_{-i}^0)  $
\[  
 = 
     \begin{cases}
       
       |\chi_h^{0}|/(N+\zeta-1) &\quad\text{if }\chi_h^{0}\in \nu^{0}_{-i} \\
       \zeta/(N+\zeta-1) &\quad\text{if } \chi_h^{0}= \emptyset \\
     \end{cases}
\]
where $\mathbf{z}_{-i}^0$ is the community indicator for all  entities excluding entity $u_i$, $\zeta$ is concentration parameter, $\nu^{0}_{-i}$ is the set $\nu^{0}$ excluding $u_i$,  $|\chi_h^{0}|$ is the number of entities in $\chi_h^{0}$ and $\emptyset$ is a new community at $t=0$. 

In the fragmentation part, each community splits into small communities and executes a CRP partition independently. The fragmentation process at $t\not=0$ is summarized as:

$\text{Frag}({z_i}^{t}):\text{p}(z_i^{t} = h|\nu^{{{t-1}'}}_{-i},\nu^{t}_{-i},z_i^{{t-1}'} =q ) = $
\[    
     \begin{cases}
       
       |\chi_h^{t}|/(|\chi_q^{{t-1}'}|+\zeta-1) &\quad\text{if }\chi_q^{{t-1}'} \in  \nu^{{t-1}'}_{-i},\chi_h^{t} \in  \nu^{{t}}_{-i}\\
       \zeta/(|\chi_q^{{t-1}'}|+\zeta-1) &\quad\text{if } \chi_q^{{t-1}'} \in  \nu^{{t-1}'}_{-i},\chi_h^{t}=\emptyset\\
       1 &\quad\text{if }\chi_q^{{t-1}'}=\emptyset,\chi_h^{t}=\emptyset\\
       0 &\quad\text{otherwise }\\
     \end{cases}
\]
We note that all the elements in $\chi_h^{t}$ also belong to $\chi_q^{{t-1}'}$.

In the coagulation part, we execute a CRP partition on the set of communities. The coagulation process at $t'$ is summarized as:

$\text{Coal}({z_i}^{t'}):\text{p}({z_i}^{t'} = e|\nu^{{t'}}_{-i},\nu^{t},{z_i}^{{t}} =h ) = $
\[    
     \begin{cases}
       
       1 &\quad\text{if }\chi_e^{{t'}} \in  \nu^{{t'}}_{-i},\chi_h^{t} \in  \nu_{-i}^{{t}}\\
       |\Omega|/(|\nu^{{t}}|+\eta-1) &\quad\text{if } \chi_e^{{t}'} \in  \nu^{{t'}}_{-i},\chi_h^{t}=\emptyset\\
       \eta/(|\nu^{{t}}|+\eta-1) &\quad\text{if }\chi_e^{{t'}}=\emptyset,\chi_h^{t}=\emptyset\\
       0 &\quad\text{otherwise }\\
     \end{cases}
\]
where $\eta$ is the concentration parameter for the coagulation process and $\Omega$ represents the communities at $t$ which belong to the community set with index $e$ at time $t'.= \{\chi_v^{t}|\chi_v^{t}	\subseteq \chi_e^{t'} \}$.

\subsection{Generating Relations}
In reality, it is common that an entity plays roles in multiple groups. For example, a doctor may be the supervisor of a nurse and the subordinate of the hospital director. Therefore, we induce MMSB to each entity at the group level by imposing a mixed membership vector $\mathbf{\theta}_i^t$ on each entity $u_i$ at a time slice $t$. ($\mathbf{\theta}_i^t $ is a membership of entity $u_i$ over $K$ groups where $\sum_k \mathbf{\theta}_i^{t,k}=1$). For each pair of entities $u_i$ and $u_j$, we sample group indicators $g_{i\rightarrow j}^t, g_{i\leftarrow j}^t$ from Multinomial($\mathbf{\theta}_i^t$) and Multinomial($\mathbf{\theta}_j^t$). The arrow in $g_{i\rightarrow j}^t$ and $g_{i\leftarrow j}^t$ indicates the sender (from $u_i$ to $u_j$) and the receiver (from $u_j$ to $u_i$), respectively.

Now, we construct a compatibility matrix to predict entity relations $x_{ij}^t$ based on the community and group indicators. Imagine that there are several communities consisting of groups inside a complex network. It is quite common that the inner structure (group relations) of each community is similar. For example, each company has sales and marketing departments. Besides, groups within the community are more likely to have tighter interactions than ones across communities. Moreover, across community, groups with similar functionality are more probable to have interactions.  Therefore two assumptions are made to construct these relations. First, group pair relations within communities are consistent. We use a compatibility matrix, $\mathbf{B}$, to model all within community group relations. Second, interactions between entities from the same group but in different communities may be  different from ones in the same group and community. To account for this we add a $K$-array across community adjustment parameter $\mathbf{Q}$ to on-diagonal values of the $\mathbf{B}$. This provides a flexible way to model the differences of within-group entity relations based on whether the entities are in the same community. Furthermore, we set the value of relations between entities that do not share community nor group to a small value, $\epsilon$. For each pair of entities $u_i$ and $u_j$, we sample $x_{ij}^t$ from Bernoulli($\frac{1}{1+\exp{({-y_{ij}^t})}}$) where
\[y_{ij}^t = 
     \begin{cases}
       {B}_{lk} &\quad\text{if } z_i^t=z_j^t \text{, }g_{i\rightarrow j}^t=l\text{, } g_{i\leftarrow j}^t=k\\
       {B}_{kk}+Q_{k}&\quad\text{if } z_i^t \neq z_j^t \text{, } g_{i\rightarrow j}^t= g_{i\leftarrow j}^t=k\\
       \epsilon &\quad\text{otherwise}\\
     \end{cases}
\]

Group pairs are always correlated in the real world. For example, employee-employer relations can be unidirectional while employee-employee may be bidirectional. 
We are interested in the correlation of group pairs so the Inverse-Wishart prior is imposed on the variance $\mathbf{\sigma}_{kl}$ of the normal distribution of ${B}_{lk}$ and ${B}_{kl}$. Finally, we share the group-level compatibility matrix $\mathbf{B}$ and adjustment parameter $K$-ary $Q$ across time due to the data sparsity.   

In summary, the fcMMSB generative model is as follows:
\begin{figure}
\centering
\includegraphics[width=0.95\columnwidth]{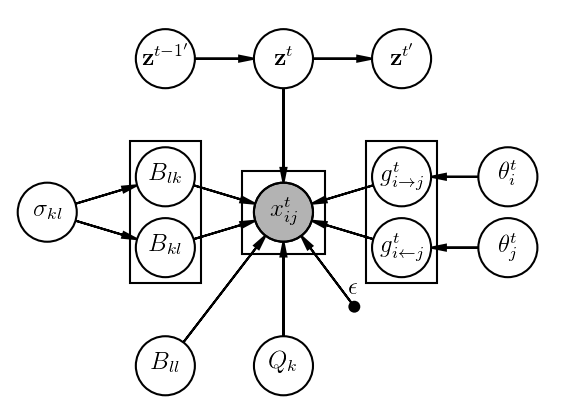}
\caption{Graphical model of fcMMSB. Hyperparameters are not shown. $\cdot^t \text{ and }\cdot^{t'}$ denote the time index of fragmentation and coagulation process respectively.  Notation: $\mathbf{z}^t = \{z_i^t|i\in\{1,...,N\}\}$.}
\label{fig:graphical model}
\end{figure}
\begin{itemize}
\item To generate compatibility matrix $\mathbf{B}$
\begin{itemize}
\item sample $\mathbf{\sigma}_{{kl}} \sim \text{Invwishart}(\varphi,\psi)$ 
\item sample $({B}_{lk},{B}_{kl}) \sim \mathcal{N}(\mathbf{\mu}_{\text{kl}},\mathbf{\sigma}_{\text{kl}})$
\item sample ${B}_{kk} \sim \mathcal{N}(\mu_{{B}},\sigma_{{B}})$ 

\end{itemize}
\item For each across community adjustment parameter $Q_{k}$
\begin{itemize}
    \item sample $Q_{k} \sim \mathcal{N}({\mu}_{\text{Q}},{\sigma}_{\text{Q}})$ 
\end{itemize}
\item For each mixed membership of entity $u_i$
\begin{itemize}
\item sample $\mathbf{\theta}_i^t \sim \text{Dirichlet}(\mathbf{\alpha})$ 
\end{itemize}
\item For each community indicator  $z_i^t$ 
\begin{itemize}
\item sample $z_i^0 \sim \text{Init}({z_i}^{0})$ 
\item sample $z_i^t \sim \text{Frag}({z_i}^{t})$ 
\item sample $z_i^{t'} \sim \text{Coal}({z_i}^{t'})$ 
\end{itemize}
\item To generate each directed relations $x_{ij}^t$
\begin{itemize}
\item sample sender group $g_{i\rightarrow j}^t \sim \text{Multinomial}(\mathbf{\theta}_i^t)$ 
\item sample receiver group $g_{i\leftarrow j}^t \sim \text{Multinomial}(\mathbf{\theta}_j^t)$ 
\item sample $x_{ij}^t \sim \text{Bernoulli}(\frac{1}{1+e^{-y_{ij}^t}})$ 
\end{itemize}
\end{itemize}
We give the graphical model of fcMMSB in Figure \ref{fig:graphical model}.

\section{Inference}
Our model is intractable for exact inference, instead we derive a Gibbs sampling scheme for posterior inference. The target is to predict the unobserved relations between entities by inferring parameters $\mathbf{z},\mathbf{\theta},\mathbf{B},\mathbf{Q},\mathbf{g} $ and $\mathbf{\sigma}$. The parameter in bold represents its total set. 
The joint distribution $\text{p}(\mathbf{x},\mathbf{z},\mathbf{\theta},\mathbf{\text{B}},\mathbf{Q},\mathbf{g}|\epsilon,\mathbf{\alpha},\zeta,\eta)$ can be expressed as:
\begin{align*}
&\prod_{i,j,t}\text{p}(x_{ij}^t|z_i^t,z_j^t,Q_{g_{i\rightarrow j}^t},\text{B}_{g_{i\rightarrow j}^t g_{i\leftarrow j}^t},\epsilon)\prod_i\text{Init}(z_i^0)\\
&\prod_{i,t}\text{Frag}(z_i^t)\text{Coal}(z_i^{t'})\prod_{k}\text{p}(Q_{k}|\mu_Q,\sigma_Q)\text{p}(\text{B}_{kk}|\mu_\text{B},\sigma_\text{B})\\
&\prod_{i,j,t}\text{p}(g_{i\rightarrow j}^t|\mathbf{\theta}_i^t)\prod_{l,k,l\neq k}\text{p}(\text{B}_{lk},\text{B}_{kl}|\mathbf{\mu}_{kl},\mathbf{\sigma}_{kl})\prod_{i,t}\text{p}(\mathbf{\theta}_i^t|\mathbf{\alpha})
\end{align*}

\subsection{Sampling $B_{lk},B_{kl} (l \neq k)$ Using Polya-Gamma }
For simplicity, the $({B}_{lk} , {B}_{kl})$ pair is denoted as a vector $\hat{\mathbf{B}}$ in this section. The Polya-Gamma~(PG) data augmentation is implemented for $\hat{\mathbf{B}}$.  Following \cite{polson2013bayesian}, $\frac{{(e^{\phi})}^m}{{(1+e^{\phi})}^n}$ can be expressed as 
$2^{-n}e^{\kappa\phi}\mathbb{E}\{e^{-w\phi^2/2}\}$ with a PG variable
$\omega \sim \text{PG}(n,0)$, where $\kappa = m-n/2$. Furthermore, with conditional distribution $p(w|\phi)$,
we have $\omega|\phi \sim \text{PG}(n,\phi)$.
Assuming that the prior of $\phi$ follows $\mathcal{N}(\mu,\sigma)$ with likelihood $\frac{{(e^{\phi})}^m}{{(1+e^{\phi})}^n}$, the posterior of $\phi$
is a Gaussian distribution. Therefore, the true posterior of $\phi$ can be derived by updating $\phi$ and $\omega$ alternately.

In our model, $\hat{\mathbf{B}}$ is updated via PG approach by alternately sampling $\hat{\mathbf{B}},\omega_{lk},\omega_{kl}$:
\begin{align*}
    &\quad \quad \qquad \hat{\mathbf{B}}|- \sim \mathcal{N}(\mu^*,\sigma^*)\\
    &\omega_{lk} \sim \text{PG}(n_{lk},B_{lk})
    , \omega_{kl} \sim \text{PG}(n_{kl},B_{kl})
\end{align*}
where
\begin{align*}
    & \mu^* = \sigma^*(\kappa + \mathbf{\sigma}_{kl} \mathbf{\mu}_{kl})\\
    & \sigma^* = {( \Omega +{\mathbf{\sigma}_{kl}}^{-1})}^{-1}
\end{align*}
$\kappa = (\kappa_{lk},\kappa_{kl})$. $\Omega$ is a diagonal matrix of $\omega_{lk}$ and $\omega_{kl}$. $\kappa_{lk} = {n_{lk}^1} - {n_{lk}}/2$. Here
$n_{lk} = \sum_{t,i,j}\mathbb{I}[g_{i\rightarrow j}^t=l] \cdot
\mathbb{I}[g_{i\leftarrow j}^t=k] \cdot \mathbb{I}[z_i^t=z_j^t]$ and $n_{lk}^1 = \sum_{t,i,j} \mathbb{I}[g_{i\rightarrow j}^t=l] \cdot
\mathbb{I}[g_{i\leftarrow j}^t=k] \cdot
\mathbb{I}[z_i^t=z_j^t] \cdot \mathbb{I}[x_{ij}^t=1]$ where $\mathbb{I}$ is an indicator function.
As the sampling scheme of $B_{ll}$ and $Q_l$ is similar with $\hat{\mathbf{B}}$, we omit the procedure here.

\subsection{Sampling $g_{i\rightarrow j}^t$}
Collapsed Gibbs sampling is used on $g_{i\rightarrow j}^t$ by marginalizing over $\mathbf{\theta}_i^t$. The posterior of $g_{i\rightarrow j}^t$ can be expressed as:
\begin{equation*}
{\text{p}(g_{i\rightarrow j}^t=k|-) \propto {\frac{{[e^{y_{ij}^t}]}^{\mathbb{I}[x_{ij}^t=1]}}{1+e^{y_{ij}^t}}}\frac{n_k^{i\neg j}(t)+\alpha_k}{\sum_k n_k^{i\neg j}(t)+\alpha_k}}     
\end{equation*}
where $n_k^{i\neg j}(t)=\sum_{l,l\neq j} \mathbb{I}[g_{i\rightarrow l}^t=k]$.

\subsection{Sampling $\mathbf{z}$ }

The prior of latent communities sequence $\mathbf{z}_i$ is:
\begin{equation*}
   \text{p}_{\text{prior}}(\mathbf{z}_i)= \text{Init}({z_i}^{0}) \cdot \text{Coal}({z_i}^{0'}) \cdot  \ldots \cdot \text{Frag}({z_i}^{T-1})
\end{equation*}  
so the posterior of $\mathbf{z}_i$ can be described as:
\begin{align*}
\text{p}(\mathbf{z}_i|-)&\propto \text{p}(\mathbf{x}_{i\cdot},\mathbf{x}_{\cdot i}|\mathbf{z},\mathbf{\theta},\text{B},\text{Q},\mathbf{g},\epsilon)\cdot \text{p}_{\text{prior}}(\mathbf{z}_i)\\
&=\prod_{j,t}\frac{{[e^{y_{ij}^t}]}^{\mathbb{I}[x_{ij}^t=1]}}{1+e^{y_{ij}^t}}\cdot \frac{{[e^{y_{ji}^t}]}^{\mathbb{I}[x_{ji}^t=1]}}{1+e^{y_{ji}^t}}\cdot \text{p}_{\text{prior}}(\mathbf{z}_i) 
\end{align*}
where $y_{ij}^t$ follows the previous definition in section $3$. For computational simplicity, we use forward-backward algorithm on $\text{p}({z}_i^{\mathbf{T}}|-)$. Here $\mathbf{x}_{i\cdot}=\{x_{ij}^t| j \in \{1,...,N\}, t \in \{0,...,{T-1}\}\}, \mathbf{x}_{\cdot i}$ is defined similarly.

\subsection{Sampling $\mathbf{\sigma}_{kl}$}

As the prior and likelihood of $\mathbf{\sigma}_{kl}$ are a conjugate pair, we give the posterior of $\mathbf{\sigma}_{kl}$ directly.
\begin{equation*}
\mathbf{\sigma}_{kl}|- \sim  \text{Invwishart}(1+\varphi,\mathbf{\psi}+(\hat{\mathbf{B}}-\mathbf{\mu}_{kl}){(\hat{\mathbf{B}}-\mathbf{\mu}_{kl})}^\intercal)     
\end{equation*}

\subsection{Prediction}
In the previous sections, we derived the samples at each iteration. We would like to use these samples to estimate the unobserved relations.  Our prediction target at iteration $s$, $\hat{x}_{ij}^{t[s]}$, is expressed as ${\hat{\mathbf{\theta}}_i^{\text{t}\intercal}} \cdot \bar{\mathbf{B}} \cdot {\hat{\mathbf{\theta}}_j^t}
$, where the superscript of $\hat{\mathbf{\theta}}_i^{\text{t}\intercal}$ is the transpose of the vector. Here each dimension of ${\mathbf{\theta}}_j^t$ is 
${\hat{\mathbf{\theta}}_i}^{t,k}=\frac{n_k^i(t)+\alpha_k}{\sum_k n_k^i(t)+\alpha_k}$ and $
 n_k^i(t) = \sum_j \mathbb{I}[g_{i\rightarrow j}^t=k]$. Each entry $\bar{\mathbf{B}}_{lk}$ of $\bar{\mathbf{B}}$ is $
 \frac{1}{1+\exp{({-\bar{\mathbf{Y}}_{lk}})}}$ and $\bar{\mathbf{Y}}_{lk}=\mathbb{I}[z_i^t=z_j^t]\mathbf{B}_{lk}+\mathbb{I}[l=k]\mathbb{I}[z_i^t \neq z_j^t](\mathbf{B}_{lk}+Q_k)+ 
\mathbb{I}[l\neq k]\mathbb{I}[z_i^t \neq z_j^t]\epsilon$.

\section{Related Work}

\begin{table*}
\centering
\begin{tabular}{l|l|l|l|l|l}
\toprule
Model    & Coleman & Student net& Mining reality & Hypertext 2009 & Infectious  \\
\midrule
CN      
& $0.881\pm 0.018$  
& $0.839\pm 0.019$
& $0.873\pm 0.004$ 
& $0.776\pm 0.006$
& $0.883\pm 0.014$
\\

MMSB      
& $ 0.880\pm 0.016$ 
& $ 0.914\pm 0.011$
& $ 0.885\pm 0.007$ 
& $ 0.867\pm 0.005$
& $ {0.965\pm 0.001}$
\\     

$\text{T-MBM}^{*}$      
& $0.881 \pm 0.005$ 
& $0.896 \pm 0.010$
& $0.861 \pm 0.002$ 
& $0.790 \pm 0.004$
& $0.838 \pm 0.008$\\

$\text{BPTF}^{*}$  
& $ 0.908 \pm 0.013$ 
& $0.909 \pm 0.021$
& $0.922 \pm 0.001$ 
&$ 0.874\pm 0.006$
& $ 0.843\pm 0.011$\\

$\text{SVD++}^{*}$  
& $ \hspace{2.2em}$ ---
& $ \hspace{2.2em}$ ---
& $ 0.833\pm 0.006$ 
& $ 0.735\pm 0.004$
& $ 0.614\pm 0.011$\\

$\text{DRGPM}^{*}$  
& $ \hspace{2.2em}$ ---
& $0.823 \pm 0.014$
& $0.933 \pm 0.003$ 
& $\mathbf{0.904 \pm 0.008 }$
& $\mathbf{0.988 \pm 0.000}$\\

MNE     
& $ 0.891\pm0.024 $ 
& $0.940 \pm 0.020$
& $0.813 \pm 0.004$  
& $ 0.872 \pm 0.008$
& $0.900 \pm 0.017$
\\

DeepWalk 
& $\mathbf{0.914\pm0.018}$  
& $0.910 \pm 0.018$
& $0.759\pm0.004$ 
& $0.816\pm0.005$
&$ 0.910\pm 0.014$
\\

\midrule
fcMMSB    
&$ 0.908 \pm 0.009$  
&$\mathbf{0.954\pm 0.006}$
&$\mathbf{0.935 \pm 0.004}$   
&${0.902 \pm 0.001}$  
&${0.981\pm 0.001}$
\\
\bottomrule
\end{tabular}
\caption{Model performance: AUC (mean and standard deviation) on the real dataset. Note: * represents a dynamic model.}
\label{tab:dynamic}
\end{table*}

The Stochastic Block Model~(SBM) presents is an earlier approach on modelling network data. In general, SBM builds on fundamental works \cite{aldous1981representations} and \cite{hoover1979relations} that bring the notion of relational data exchangeability~\cite{AISTAS2018_BSPT,NIPS2018_RBP}. Blockmodels \cite{dyer1989solution} and \cite{snijders1997estimation} leverage interactions between entities to generate corresponding clusters that have a symmetric adjacency matrix. The Infinite Relational Model (IRM) \cite{kemp2006learning} makes an extension to SBM by allowing the number of clusters to be undetermined. Our proposed work is based on MMSB, the key contribution of which is to allow each entity to hold multiple groups in a network. 

Another important class of relation model, Poisson matrix factorization model, is also used in the evaluation section. Bayesian Poisson Tensor Factorization~(BPTF) \cite{schein2015bayesian} models the dyadic events via a tensor factorization. \cite{zhou2015infinite} focuses on utilizing  hierarchical
gamma process on static networks mainly. \cite{yang2018dependent,yang2018poisson} make substantial contributions of incorporating the completely random measures into the modelling, and \cite{NIPS2019_SDREM} proposes a deep and scalable version of the Mixed-Membership Stochastic Blockmodel.

DFCP provides flexibility to model the process of splitting and merging of communities. The predecessor to DFCP, the fragmentation coagulation process (FCP) \cite{teh2011modelling} is a continuous limit of DFCP. The main drawback of FCP is that at most two clusters can undergo a merge operation or one cluster can be split into at most two clusters at the same time. There is no constraint on the number of clusters to be split or merged in DFCP.

We derive a highly efficient sampling scheme via a data augmentation approach (PG) \cite{polson2013bayesian}. This approach offers a model that has a Bernoulli likelihood with a Gaussian prior transferred by the logistic function. Naturally, \cite{durante2014bayesian} implements PG approach on Gaussian process (GP). Furthermore, \cite{zhou2012lognormal} improves Poisson regression by LGNB model with PG approach.

\section{Evaluation}

\subsection{Synthetic Data}
\begin{figure}
\centering
\includegraphics[width=.95\columnwidth]{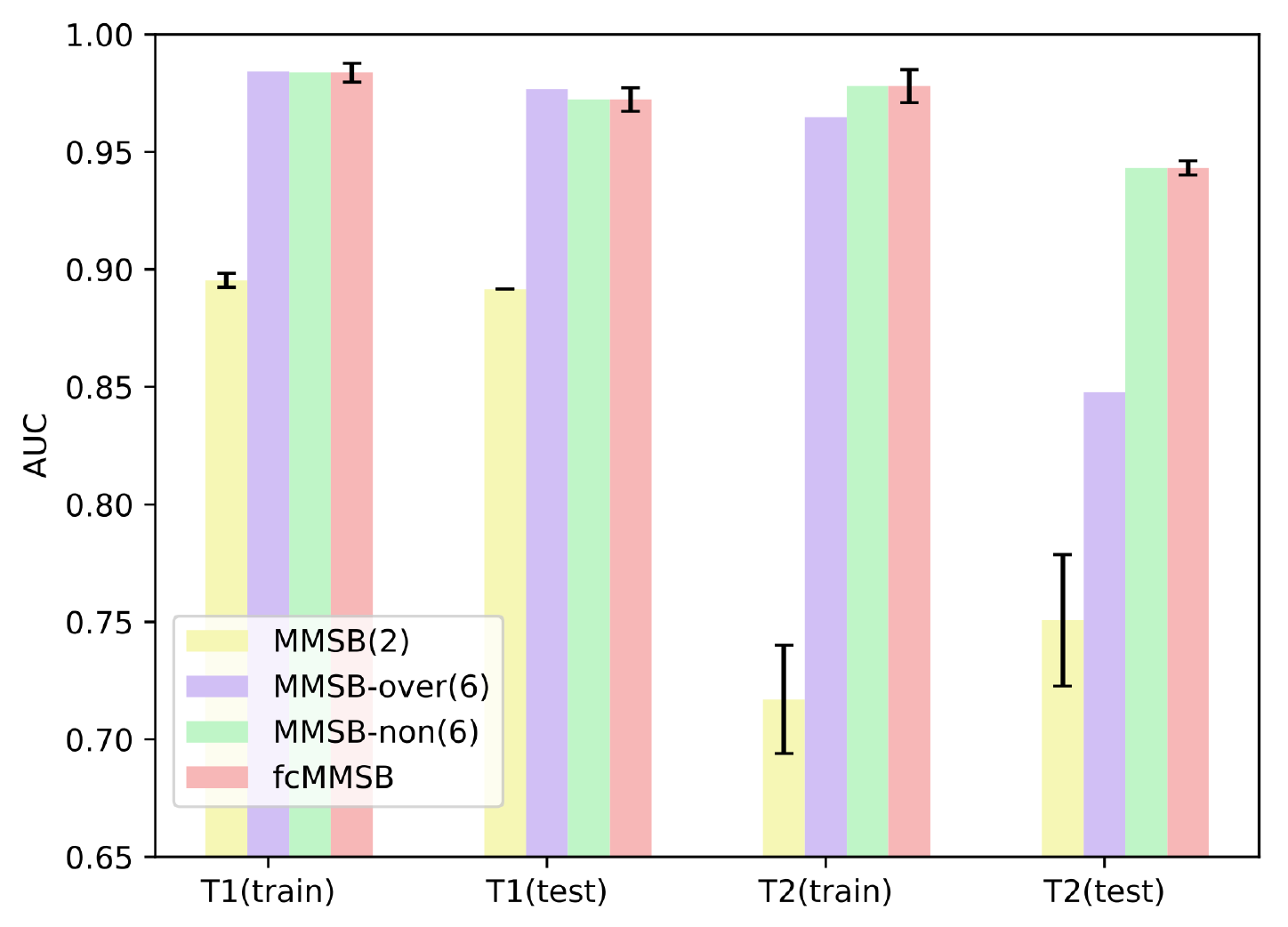}
\caption{AUC comparison on synthetic data.}
\label{fig:syn}
\end{figure}
To demonstrate the problem of the MMSB mentioned in the introduction, we generate a synthetic dataset with $N=100$ and $T=2$, the generative process for which is described as follows:

\begin{enumerate}
    \item Instantiate a network structure of three communities containing two groups each. For each time slice, generate the mixed membership for 100 entities by sampling the Dirichlet distribution with parameters $[0.8,0.2] \text{ or } [0.2,0.8]$ depending on the group. Set $\mathbf{B}$ to be a $2 \times 2$ compatibility matrix with high on-diagonal values and low off-diagonal values.
    \item For time slice 1, if both entities belong to the same community perform step 3, otherwise set the entity relation to 0. For time slice 2, if both entities belong to the same community and group perform step 3, otherwise set the entity relation to 0.
    \item Generate entity relations using the Bernoulli distribution with parameter ($\mathbf{\theta}^\intercal_i\mathbf{B}\mathbf{\theta}_j$) for the relation between $u_i$ and $u_j$.
\end{enumerate}

For evaluation, we randomly split the data into 2 subsets: $80\% $ for training and $20\%$ for testing. We compare our model with two different MMSB models varying in the number of groups in the compatibility matrix. The train and test AUC results are provided in Figure \ref{fig:syn}. We notice that when the number of groups in MMSB is 2, it is underfitting relative to fcMMSB with 2 groups. When the number of groups in MMSB is 6, there are two possible outcomes: overfitting and not overfitting. The overfitting of the MMSB is demonstrated by the higher train AUC and lower test AUC on time slice 2 compared to our model. Overfitting is not always the outcome, however, and the stochastic nature of the MMSB means that on different runs, the MMSB may achieve similar results to our model, as shown by MMSB-non in Figure \ref{fig:syn}. This demonstrates the problem of choosing the number of groups in the MMSB.

\begin{figure}
\centering
\includegraphics[width=.95\columnwidth]{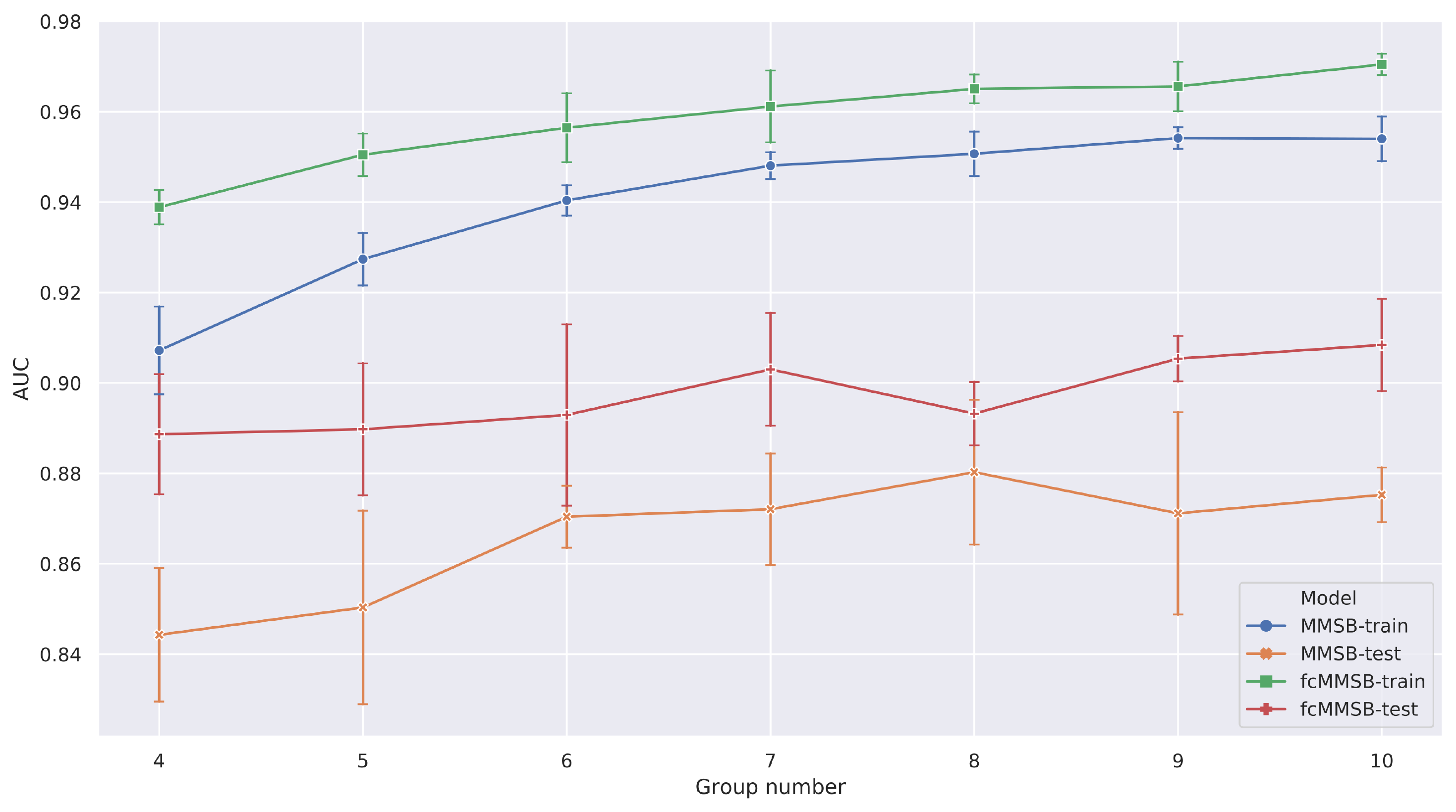}
\caption{
Comparison of AUC between MMSB and fcMMSB on the Coleman dataset.}
\label{fig:auc comparison on coleman dataset}
\end{figure}

\subsection{Prediction Relations}
To demonstrate the potential of our fcMMSB model, we use five real-world datasets for validation. We use the relation prediction task to validate our model.
The area under the ROC (Receiver Operating Characteristic) curve (AUC) is used as a performance metric. Here, we randomly select 80\% data for training and leave the 20\% for testing. 
Each experiment is run for five times, and we report the AUC results with their mean and standard deviation values. Five real-world datasets are described as follows:

\begin{itemize}
    \item The Coleman dataset \cite{coleman1964introduction} contains the information about the friendships of boys in an Illinois high-school. It records the three closest friends for each student in the fall of 1957 and spring of 1958. The binarized dataset is a $73 \times 73 \times 2$ asymmetric matrix. 
    \item The Student net dataset \cite{fan2014dynamic} describes the relations between students. We binarize the relations at each time slice, leading to a $50 \times 50 \times 3$ asymmetric matrix.
    \item Mining Reality dataset \cite{eagle2006reality} records contact data of 96 students at the Massachusetts Institute of Technology (MIT) over 9 months in 2004. The dataset is split into 10 time slices, then we set each entity pair value to be 1 at that time slice if they have at least one contact during that time. Thus, it leads to a $96 \times 96 \times 10$ symmetric matrix.
    \item The Hypertext 2009 dataset \cite{isella2011s} records the contact network  ACM Hypertext 2009 conference attendees. The relation between two attendees is 1 if they have a face-to-face contact over 20 seconds. We split the dataset into 10 time slices and binarize it, leading to $113 \times 113\times 10$ symmetric matrix.
    \item The Infectious dataset \cite{isella2011s} describes the face-to-face interactions between people during the exhibition INFECTIOUS: STAY AWAY in 2009 at the Science Gallery in Dublin. Each relation is 1 if those two people had face-to-face contact for at least 20 seconds. We binarize the relations at each time slice, leading to a $410 \times 410 \times 10$ symmetric matrix.
   
\end{itemize}

\begin{figure}

\centering
\includegraphics[width=.95\columnwidth]{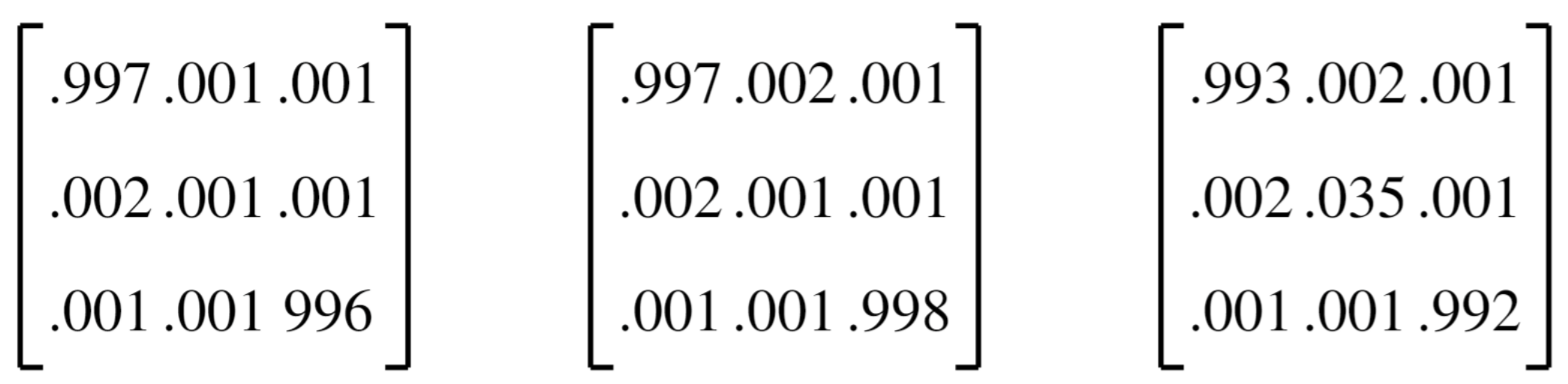}
\caption{Left: compatibility matrix in MMSB. Middle: compatibility matrix within community in fcMMSB. Right: compatibility matrix across community in fcMMSB.}
\label{fig:compatibility matrix}
\end{figure}

\begin{figure}

\centering
\includegraphics[width=1\columnwidth]{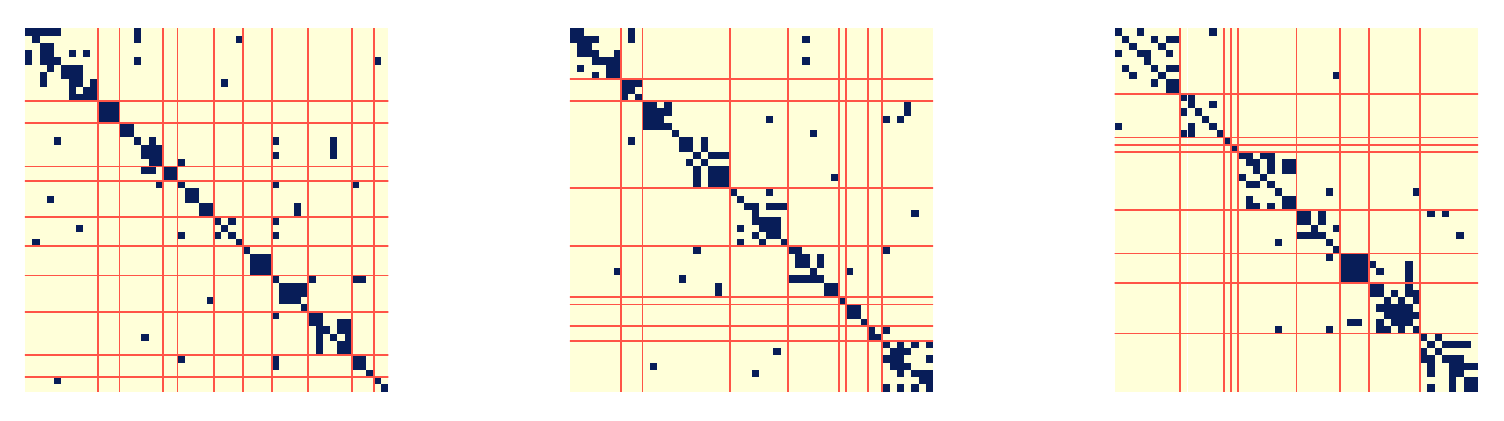}
 
\caption{
Visualization of community clustering on the Student net dataset across time.}
\label{fig:stu_net}
\end{figure}

\begin{figure}

\centering
\includegraphics[width=1\columnwidth]{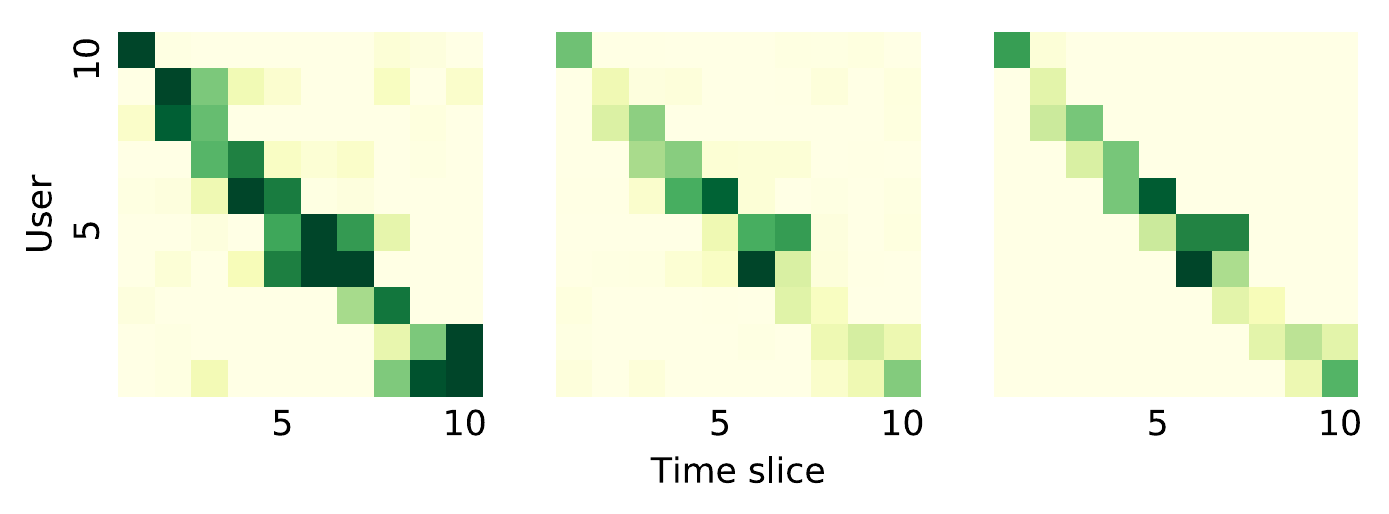}
\caption{
Left: User activeness across time. Middle: The recovered number of user interactions. Right: The original number of user interactions.}
\label{fig:infectious}
\end{figure}

\subsection{General Performance}
We use eight baseline methods for comparison. 
One structure-based model: Common neighbor (CN) \cite{newman2001clustering}. Five feature or cluster based models: Mixed Membership Stochastic Blockmodel (MMSB) with Gibbs sampling \cite{airoldi2008mixed}, Temporal Tensorial Mixed Membership Stochastic Blockmodel  (T-MBM) \cite{tarres2019tensorial}, Bayesian Poisson Tensor Factorization (BPTF) \cite{schein2015bayesian}, Collaborative filtering with temporal dynamics~(SVD++) \cite{koren2009collaborative} and Dependent relational gamma process model~(DRGPM) \cite{yang2018dependent}. 
Two embedding based models: Scalable Multiplex Network Embedding (MNE) \cite{Zhang2018ScalableMN} and  DeepWalk \cite{perozzi2014deepwalk}.

We show the results in Table \ref{tab:dynamic}. The overall result of fcMMSB is competitive with DRGPM and outperforms the other state-of-the-art models. 
This may result from fcMMSB, with its flexible structure, being more suitable for long time series datasets in which the number of communities may vary across time. The DRGPM performance on Student net dataset may suffer from the short time sequence of the dataset. 

Compared with vanilla MMSB, fcMMSB also shows its advantage on both short and long time series dataset. We compare our model with MMSB by varying the group number parameter on the Coleman dataset  in Figure \ref{fig:auc comparison on coleman dataset}. fcMMSB achieved better AUC on both train and test sets. When we increased the group number, train AUC on both models increased. Due to the flexible structure of fcMMSB, the margin of train and test AUC between fcMMSB  and MMSB is relatively larger with smaller group numbers. While the train AUC of MMSB is relatively close to that of fcMMSB, the test AUC is lower. 

Besides, we compare fcMMSB with vanilla MMSB by looking at the trained compatibility matrix for the Hypertext dataset in Figure \ref{fig:compatibility matrix}. We see that the MMSB compatibility matrix is similar to the within-community matrix in fcMMSB. However, there is a moderate difference in fcMMSB between the within-community and across communities matrices for the entry (2,2). This shows that the group-pair relation within a community is not same as the one across communities, therefore MMSB with its single compatibility matrix, cannot properly model this network. 
This is why the fcMMSB is better than MMSB on the Hypertext dataset; its more flexible structure is better at modeling multiple communities. In comparing the compatibility matrices of other datasets, we find this to be the case with other datasets as well. We also observe that the second role of the membership covers the main part for most people due to sparsity which can be interpreted as the inactive role. This interpretation is consistent with the compatibility matrix. 

In Figure \ref{fig:stu_net}, we visualize the community clustering result on the Student net dataset. We find the data points are dense along the diagonal. This is consistent with our assumption that the interactions within the community are tighter than the ones across communities. Besides, we find that most entities belong to the same communities across time, even though the community index may change. This shows why  DFCP is used in our model since DFCP constructs a temporal dependency for communities across time.

Furthermore, to show the dynamic of membership in the Infectious dataset, for each time slice, we randomly select one user who is active at that time slice. To show the intensity of user activeness, we define activeness, $\text{AC}$, for each user $i$ at time slice $t$ to be $\text{AC}_i^t = {\theta_i^t}^{\intercal}\cdot \mathbf{B}\cdot \mathbf{1}$. We present the user activeness and the recovered number of users' interactions with the original one in Figure \ref{fig:infectious}. It is interesting that the user is active in consecutive time segments. Meanwhile, comparing the user activeness with the original user interactions, it is easy to observe that they have correlations. This shows the membership used for user activeness really reflects the characteristic of the data. Also the recovered number of user interactions is similar with the original one in Figure \ref{fig:infectious}.  Besides, we find that BPTF got the relatively low AUC compared with the other four datasets. It seems that the tight correlation of features across time inherent in BPTF does not fit this dataset. Overall, fcMMSB is stable in both dense (Coleman, Student net, Mining reality) and sparse (Hypertext, Infectious) datasets.

\section{Conclusion}
In this work, we highlight two problems in MMSB: the structure in MMSB is unable to encapsulate the prior information like the community structure of entities in the static case; and modelling the community evolution using a fixed size compatibility matrix may suffer underfitting/overfitting in the dynamic case. To overcome these two problems, we developed the fragmentation coagulation based Mixed Membership Stochastic Blockmodel (fcMMSB). Specifically, we used CRP for entity-based clustering to capture the community information of entities and MMSB for linkage-based clustering to derive the group information for links simultaneously. Besides, we utilized DFCP to infer the community structure (including the number of communities) among entities and evolution (appearance/disappearance or split/merge). Our model combines a group-level compatibility matrix with a community adjustment parameter to satisfy the four types of entity pair relations: within and across communities and groups. 
Our model unifies these techniques to derive a generalized MMSB. Furthermore, a PG approach is implemented for an efficient sampling scheme to infer hidden variables. Finally, we demonstrate the fcMMSB outperforms and is competitive with the state-of-the-art methods through experiments on real datasets.

\bibliography{fcmmsb}
\bibliographystyle{aaai}

\end{document}